\pdfoutput=1

\documentclass[11pt]{article}

\usepackage{EMNLP2023}

\usepackage{times}
\usepackage{latexsym}

\usepackage[T1]{fontenc}

\usepackage[utf8]{inputenc}

\usepackage{microtype}

\usepackage{inconsolata}

\usepackage{graphicx}
\usepackage{amsmath,amsfonts,amssymb}

\title{Strong and Efficient Baselines for \\ Open Domain Conversational Question Answering}

\author{
Andrei C. Coman\thanks{~~Work was done during an internship at Amazon Science.}\\
Idiap Research Institute, EPFL \\
\href{mailto:andrei.coman@idiap.ch}{andrei.coman@idiap.ch} \\
\And
Gianni Barlacchi \\
Amazon Alexa AI \\
\href{mailto:gbarlac@amazon.com}{gbarlac@amazon.com}
\And
Adrià de Gispert \\
Amazon Alexa AI \\
\href{mailto:agispert@amazon.com}{agispert@amazon.com}
}

\begin{document}
\maketitle
\begin{abstract}
Unlike the Open Domain Question Answering (ODQA) setting, the conversational (ODConvQA) domain has received limited attention when it comes to reevaluating baselines for both efficiency and effectiveness. In this paper, we study the State-of-the-Art (SotA) Dense Passage Retrieval (DPR) retriever and Fusion-in-Decoder (FiD) reader pipeline, and show that it significantly underperforms when applied to ODConvQA tasks due to various limitations. We then propose and evaluate strong yet simple and efficient baselines, by introducing a fast reranking component between the retriever and the reader, and by performing targeted finetuning steps. Experiments on two ODConvQA tasks, namely \mbox{\textsc{TopiOCQA}} and \mbox{OR-QuAC}, show that our method improves the SotA results, while reducing reader's latency by 60\%. Finally, we provide new and valuable insights into the development of challenging baselines that serve as a reference for future, more intricate approaches, including those that leverage Large Language Models (LLMs).
\end{abstract}

\section{Introduction}
\label{sec:introduction}

\begin{figure*}
    \centering
    \includegraphics[width=\textwidth]{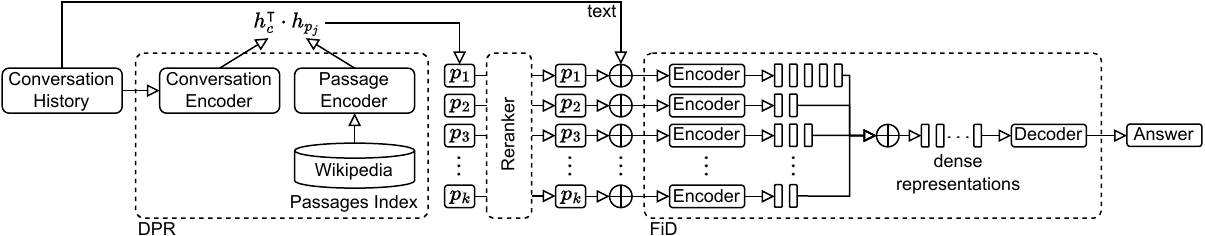}
    \caption{The \underline{R}etriever-\underline{R}eranker-\underline{R}eader (R3) pipeline}
    \label{fig:retriever_reranker_reader_pipeline}
\end{figure*}

In an automated information-seeking conversation scenario between two parties, the human questioner asks a series of questions and expects to receive a relevant response from the answering system \citep{Oddy1977INFORMATIONRT}. Current State-of-the-Art (SotA) shapes the answerer via two neural models, the Dense Passage Retrieval (DPR) \citep{karpukhin-etal-2020-dense} and the Fusion-in-Decoder (FiD) \citep{izacard-grave-2021-leveraging}, which act as retriever and reader, respectively. Their success stems from the ability to overcome certain limitations of their sparse and extractive counterparts, such as not relying on lexical retrieval heuristics or extracting spans as a response \citep{Chen2017ReadingWT, Yang2019EndtoEndOQ, lee-etal-2019-latent, 48303, pmlr-v119-guu20a, NEURIPS2020_6b493230, shen2023neural}. Among the most promising approaches are those concerning the improvement of training strategies \citep{pmlr-v119-guu20a, balachandran-etal-2021-simple, qu-etal-2021-rocketqa}, use of rerankers \citep{Hu2019RetrieveRR, mao-etal-2021-reader, barlacchi2022focusqa, iyer-etal-2021-reconsider, glass-etal-2022-re2g}, question rewriting \citep{vakulenko2021question, del2021question}, reader to retriever knowledge distillation \citep{izacard2021distilling}, memory-efficient pipeline \citep{Izacard2020AME, del2022rewriting}, and leveraging structured information \citep{Min2019KnowledgeGT, yu-etal-2022-kg}.
\newline \indent Unlike the Open Domain Question Answering (ODQA) setting, a reassessment of the baselines in terms of both efficiency and effectiveness appears to be under-explored in the conversational (ODConvQA) domain. In this paper, we focus on the typical DPR retriever and FiD reader (DPR+FiD) pipeline, and show its limitations when applied to the ODConvQA setting. Despite its popularity, we find that this baseline significantly underperforms when finetuned on downstream tasks. We show that simple improvements in the training, architecture, and inference setups of the DPR+FiD pipeline, provide a strong and efficient baseline that exceeds the performance of SotA models on two common ODConvQA datasets: TopiOCQA \citep{adlakha-etal-2022-topiocqa} and ORConvQA (\mbox{OR-QuAC}) \citep{Qu2020OpenRetrievalCQ}.
\newline \indent We point out several limitations of the pipeline, such as: 1) \textit{reader's susceptibility to noisy input}, 2) \textit{retriever's reduced coverage}, 3) \textit{retriever's lack of cross semantic encoding between the conversation and the retrieved passages}, and 4) \textit{reader's latency is heavily impacted by the number of input passages}. To mitigate these, we propose and evaluate a simple and effective approach by including a fast reranking component between the retriever and the reader, and by performing targeted finetuning steps. The proposed \underline{R}etriever-\underline{R}eranker-\underline{R}eader \underline{fine}tuning (R3FINE) strategy leads to baseline models with a better latency/performance trade-off. These baselines, which are simple and easy to replicate, serve as a reference point for comparing new and more complex models, and determining their effectiveness. Our contributions are the following:
\begin{itemize}
  \item We identify and address several limitations of the typical pipeline used in ODConvQA.
  \item We propose the R3FINE strategy, which improves SotA results on two common datasets and reduces pipeline's latency by 60\%.
  \item We provide new and valuable insights for creating simple and efficient baselines, which serve as a reference point for future comparison of new more complex approaches.
\end{itemize}
\section{End-to-End Baselines for ODConvQA}
\label{sec:end_to_end_baseline_for_odconvqa}
This section provides a brief introduction to the pipeline on which this work focuses. Figure \ref{fig:retriever_reranker_reader_pipeline} shows the typical pipeline used within the ODConvQA setting, featuring an additional reranker component. A conversation history is input to the DPR retriever. This module exploits a dual-encoder based on the BERT \citep{devlin-etal-2019-bert} model. First, it encodes the conversation history via the $ConversationEncoder$ component, which takes as input the text of the conversation history $c_{1}, c_{2}, \dots, c_{i}$, and then it outputs a dense representation $h_{c}$. Next, this representation is used to perform a dense search to retrieve the most relevant passages, i.e., text blocks that serve as basic retrieval units, from an external knowledge source (e.g., Wikipedia). The latter contains dense representations of the passages that have been encoded via the $PassageEncoder$ component, which takes as input a $j$-th passage with a given text length $N$, i.e., $p_{j_1}, p_{j_2}, \dots, p_{j_N}$, and outputs a dense representation $h_{p_j}$. The dense search is performed via the Maximum Inner-Product Search (MIPS) function which outputs the value corresponding to $h_{c}^{\intercal} \cdot h_{p_j}$.
\newline \indent Once the top-$k$ relevant passages have been retrieved, their text is appended to the conversation history and subsequently passed to the FiD reader, which is based on the T5 \citep{JMLR:v21:20-074} model. The newly created textual sequences of length $S$ are then encoded in parallel via the $Encoder$ component that outputs a dense representation $\bar{h} = \{h_1, \dots, h_S\}$. As a final step, the dense representations of the entire list of $k$ input passages are concatenated to form a single $\bar{h}_{1} \oplus \bar{h}_{2}, \dots, \oplus \bar{h}_{k}$ sequence that forms the input to the $Decoder$ component responsible for generating the answer $a$.

\section{Strong Baseline Models}
\label{sec:limitations_of_dpr_fid}

This work focuses on two main datasets. \mbox{\mbox{\textsc{TopiOCQA}}} \citep{adlakha-etal-2022-topiocqa} is a large-scale open-domain information-seeking conversational dataset that contains a challenging phenomenon in the form of topic switching. \mbox{OR-QuAC} \citep{Qu2020OpenRetrievalCQ} leverages CANARD's \citep{elgohary-etal-2019-unpack} context-independent question rewrites of the QuAC \citep{choi-etal-2018-quac} dataset, and adapts it to the open-domain setting. Further details regarding the datasets are provided in Appendix \ref{sec:appendix}.
\newline \indent We outline a number of limitations of the typical DPR+FiD pipeline, along with suggestions on how to mitigate them. While some of those interconnect at different levels the various efforts made in the ODQA domain \citep{balachandran-etal-2021-simple, yu-etal-2022-kg}, our goal is to offer a perspective on the ODConvQA setting.

\subsection{Current Limitations and Bottlenecks}
\label{subsec:current_limitations_and_bottlenecks}

\paragraph{Reader's susceptibility to noisy input.} Previous findings have shown that the FiD reader performance significantly improves when increasing the number of retrieved passages \citep{izacard-grave-2021-leveraging}. While confirming this finding, in Table \ref{tab:topiocqa_performance} we also present a different perspective to it. We show that when the same reader model is provided with the relevant (i.e., gold) passage in input, the performance decreases as the number of retrieved passages increases. This suggests that there is a balance in presenting input to the reader: if the gold passage is present, i.e., the retriever could retrieve it, a small relevant list is best, but otherwise a larger list is better.

\begin{table}
\centering
\begin{tabular}{c| c c| c c}
    \hline
    \multicolumn{5}{c}{\textbf{\mbox{\textsc{TopiOCQA}}}} \\
    \hline
    {} & \multicolumn{2}{c|}{\textbf{w/o gold}} & \multicolumn{2}{c}{\textbf{w/ gold}} \\
    \hline
    \multicolumn{1}{c|}{\textbf{top-$k$}} & \multicolumn{1}{c|}{\textbf{EM}} & \multicolumn{1}{c|}{\textbf{F1}} & \multicolumn{1}{c|}{\textbf{EM}} & \multicolumn{1}{c}{\textbf{F1}} \\
    \hline
    1 & 19.3 & 37.6 & 38.3 & 65.5 \\
    10 & 29.8 & 52.4 & 35.8 & 61.5 \\
    50 & 33.0 & 55.1 & 35.9 & 59.5  \\
    \hline
\end{tabular}
\caption{FiD reader performance (Exact Match and F1 scores) on the \mbox{\textsc{TopiOCQA}} dev split, with/without the gold passage (w/o gold) in the top-$k$ limit.}
\label{tab:topiocqa_performance}
\end{table}

\paragraph{Retriever's reduced coverage.} Current solutions impose a hard top-$k$ limit on the number of passages returned by the DPR retriever and assume that the relevant ones are present within this limit. Table \ref{tab:topiocqa_performance} shows that coverage is key during the retrieval phase for the reader to perform well. To improve it, we suggest introducing a simple and efficient Transformer-based \citep{Vaswani2017AttentionIA} reranker component after the retriever. This component, shown in Figure \ref{fig:retriever_reranker_reader_pipeline} and described in the next paragraph, is designed to reconsider a larger pool of passages returned by the DPR and to provide the FiD with a reduced and improved list of passages. Since this module operates at the semantic level, we refer to it as the $SemanticReranker$. Table \ref{tab:topiocqa_orquac_dev_coverage} shows the potential coverage margins and the retrieval results obtained after the introduction of such a module, when a larger number of passages (50 vs 1000) is considered.

\paragraph{Retriever's lack of cross semantic encoding between the conversation and the retrieved passages.} The DPR retriever performs independent encoding of the passages via the $PassageEncoder$ function. This means that it is not able to exploit the semantic relationship among them. This can be mitigated via the introduction of the previously mentioned $SemanticReranker$ component. This new module is based on the \textit{TransformerEncoder} and applies the following function:
\[
\tilde{h}_c, \tilde{h}_{p_1}, \dots, \tilde{h}_{p_k} = \\
Reranker(h_c, h_{p_1}, \dots, h_{p_k})
\]
where each element of the input attends to both the conversation dense representation $h_c$ and passages dense representations $h_{p_i}$. Reranking is performed over the new output sequence $\tilde{h}_c, \tilde{h}_{p_1}, \dots, \tilde{h}_{p_k}$ via the previously mentioned MIPS function.

\begin{figure}[ht]
    \centering
    \includegraphics[width=0.48\textwidth]{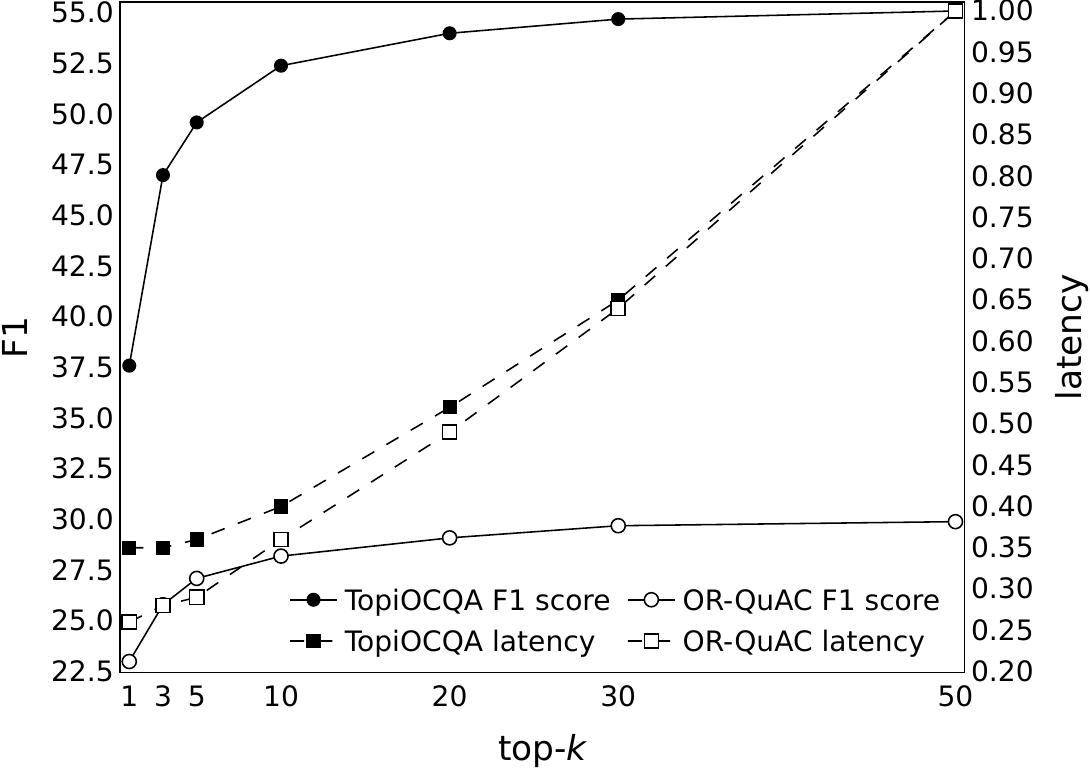}
    \caption{FiD reader performance (F1 score and latency) on the \mbox{\textsc{TopiOCQA}} dev split and \mbox{OR-QuAC} test split, with varying top-$k$ input passages. Latency is relative to the top-$50$ (top-$k$ vs top-$50$).}
    \label{fig:latency_vs_f1_score}
\end{figure}

\paragraph{Reader's latency is heavily impacted by the number of input passages.} Figure \ref{fig:latency_vs_f1_score} shows that the latency of the reader can be significantly reduced by decreasing the number of input passages. However, a trivial limitation to top-$k$ considerably degrades the performance of the module, thus leading to an inevitable trade-off. The task of the $SemanticReranker$ involves pushing relevant passages into the top-$k$ list, and allowing for a low $k$ value to be set.

\begin{table}[!t]
\centering
\begin{tabular}{c| c c| c c}
    \hline
    \multicolumn{1}{c|}{} & \multicolumn{2}{c|}{\textbf{\mbox{\textsc{TopiOCQA}}}} & \multicolumn{2}{c}{\textbf{\mbox{OR-QuAC}}} \\
    \hline
    \multicolumn{1}{c|}{\textbf{top-$k$}} & \multicolumn{1}{c|}{\textbf{w/o SR}} & \multicolumn{1}{c|}{\textbf{w/ SR}} & \multicolumn{1}{c|}{\textbf{w/o SR}} & \multicolumn{1}{c}{\textbf{w/ SR}} \\
    \hline
    1 & 24.66 & 42.64 & 29.27 & 56.64 \\
    10 & 62.45 & 75.62 & 57.14 & 72.27 \\
    50 & 77.41 & 84.69 & 65.22 & 73.53 \\
    \hline
    1000 & 91.49 & 91.49 & 74.55 & 74.55 \\
    \hline
\end{tabular}
\caption{Dev split retrieval coverage before/after the introduction of the $SemanticReranker$ (w/o SR) when a larger number of passages is considered (50 vs 1000).}
\label{tab:topiocqa_orquac_dev_coverage}
\end{table}

\begin{table*}[!thb]
\centering
\begin{tabular}{c| c c| c c| c c| c c| c c| c c}
    \hline
    \multicolumn{1}{c|}{} & \multicolumn{6}{c|}{\textbf{\mbox{\textsc{TopiOCQA}}}} & \multicolumn{6}{c}{\textbf{\mbox{OR-QuAC}}} \\
    \hline
    \multicolumn{1}{c|}{} & \multicolumn{2}{c|}{\textbf{w/o SR}} & \multicolumn{2}{c|}{\textbf{w/ SR}} & \multicolumn{2}{c|}{\textbf{w/ SR + FT}} &  \multicolumn{2}{c|}{\textbf{w/o SR}} & \multicolumn{2}{c|}{\textbf{w/ SR}} & \multicolumn{2}{c}{\textbf{w/ SR + FT}} \\
    \hline
    \multicolumn{1}{c|}{\textbf{top-$k$}} & \multicolumn{1}{c|}{\textbf{EM}} & \multicolumn{1}{c|}{\textbf{F1}} & \multicolumn{1}{c|}{\textbf{EM}} & \multicolumn{1}{c|}{\textbf{F1}} & \multicolumn{1}{c|}{\textbf{EM}} & \multicolumn{1}{c|}{\textbf{F1}} & \multicolumn{1}{c|}{\textbf{EM}} & \multicolumn{1}{c|}{\textbf{F1}} & \multicolumn{1}{c|}{\textbf{EM}} & \multicolumn{1}{c|}{\textbf{F1}} & \multicolumn{1}{c|}{\textbf{EM}} & \multicolumn{1}{c}{\textbf{F1}} \\
    \hline
    1 & 19.3 & 37.6 & 28.1 & 50.4 & 30.7 & 52.4 & 14.7 & 23.0 & 13.9 & 25.7 & 16.6 & 28.9 \\
    10 & 29.8 & 52.4 & 33.2 & 57.3 & \textbf{35.8} & \textbf{59.0} & 19.0 & 28.2 & 19.4 & 30.0 & 21.6 & \textbf{32.9} \\
    50 & \underline{33.0} & \underline{55.1} & 33.9 & 56.2 & 35.2 & 56.1 & \underline{22.0} & \underline{29.9} & 22.1 & 30.2 & \textbf{23.5} & 32.2 \\
    \hline
\end{tabular}
\caption{FiD reader performance (Exact Match and F1 scores) on the \mbox{\textsc{TopiOCQA}} dev split and \mbox{OR-QuAC} test split before/after the introduction of the $SemanticReranker$ (w/o SR), together with the results obtained after a further reader finetuning step with top-$10$ output by the SR (w/ SR + FT). \underline{Underlined} values indicate the results obtained by the DPR+FiD pipeline. \textbf{Bold} values indicate the results obtained after the introduction of the SR together with targeted finetuning steps.}
\label{tab:reader_results}
\end{table*}

\begin{table}[!thb]
\centering
\begin{tabular}{l| c c}
    \hline
    \multicolumn{3}{c}{\textbf{\mbox{\textsc{TopiOCQA}}}} \\
    \hline
    \multicolumn{1}{l|}{\textbf{Model}} & \multicolumn{1}{c|}{\textbf{EM}} & \multicolumn{1}{c}{\textbf{F1}} \\
    \hline
    BM25 + DPR Reader & 13.6 & 25.0 \\
    BM25 + FiD & 24.1 & 37.2 \\
    DPR Retriever + DPR Reader & 21.0 & 43.4 \\
    DPR Retriever + FiD & 33.0 & 55.3 \\
    \hline
    (Ours) DPR Retriever + FiD & 33.0 & 55.1 \\
    + R3FINE (top-$10$) & \textbf{35.8} & \textbf{59.0} \\
    \hline
\end{tabular}
\caption{\mbox{\textsc{TopiOCQA}} dev split baselines performance (Exact Match and F1 scores) comparison between sparse/dense retrievers (BM25/DPR Retriever) and extractive/generative readers (DPR Reader/FiD).}
\label{tab:topiocqa_baselines_comparison}
\end{table}

\subsection{A Strong and Efficient Baseline}
\label{subsec:a_strong_and_efficient_baseline}
Based on the findings above, we introduce the \underline{R}etriever-\underline{R}eranker-\underline{R}eader \underline{fine}tuning (R3FINE) strategy, which can be used to design strong and efficient baselines for ODConvQA. First, we increase the number of passages returned by the DPR from the initial 50 to 1000. Then, we add the $SemanticReranker$ component, which corresponds to a single $TransformerEncoder$ layer. We train/finetune the $SemanticReranker$ along with the $ConversationEncoder$ while keeping the $PassageEncoder$ frozen. Guided by the intuition that less but more relevant passages are beneficial to FiD as reported in Table \ref{tab:topiocqa_performance}, we finally perform an additional finetuning step by leveraging the new top-$10$ list of passages returned by the $SemanticReranker$.

\section{Experiments and Results}
\label{sec:experiments_and_results}

This section shows the impact that the introduction of the $SemanticReranker$ module has on the pipeline, as well as the finetuning steps we followed to make the pipeline more efficient without compromising its performance.

\paragraph{Experimental Setup.} As the starting point of our experiments, we used the DPR and FiD models provided with the \mbox{\textsc{TopiOCQA}} dataset. Currently, only the \textit{train} and \textit{dev} splits are made available for this dataset. We followed the same experimental setup and exploited \mbox{\textsc{TopiOCQA}}'s DPR module for both datasets. Unlike \mbox{\textsc{TopiOCQA}}, \mbox{OR-QuAC} is of extractive type, and for this reason we trained the FiD module from scratch by following the same training configuration of \mbox{\textsc{TopiOCQA}}. 

\begin{table}
\centering
\begin{tabular}{l| c c}
    \hline
    \multicolumn{3}{c}{\textbf{\mbox{OR-QuAC}}} \\
    \hline
    \multicolumn{1}{l|}{\textbf{Model}} & \multicolumn{1}{c|}{\textbf{EM}} & \multicolumn{1}{c}{\textbf{F1}} \\
    \hline
    DrQA \citep{Chen2017ReadingWT} & - & 6.3 \\
    BERTserini \citep{Yang2019EndtoEndOQ} & - & 26.0 \\
    ORConvQA \citep{Qu2020OpenRetrievalCQ} & - & 29.4 \\
    \hline
    (Ours) DPR Retriever + FiD & \textbf{22.0} & 29.9 \\
    + R3FINE (top-$10$) & 21.6 & \textbf{32.9} \\
    \hline
\end{tabular}
\caption{\mbox{OR-QuAC} test split baselines performance (Exact Match and F1 scores) comparison.}
\label{tab:orquac_baselines_comparison}
\end{table}

\paragraph{End-to-End Results.} Table \ref{tab:topiocqa_baselines_comparison} and Table \ref{tab:orquac_baselines_comparison} compare our R3FINE strategy with previous baselines. R3FINE achieves an F1 score of 59 points on \mbox{\textsc{TopiOCQA}}, and 32.9 on \mbox{OR-QuAC}, which are 3.9 and 3 points higher than the best models proposed in the original papers. It is worth noting that these large improvements are achieved with simple adjustments in the training, architecture, and inference setups of the well-established DPR+FiD baseline, and not via the introduction of new heavier and complex models.
\newline \indent To further support our R3FINE strategy, in Table \ref{tab:reader_results} we present an ablation study which quantifies its impact on the DPR+FiD pipeline. We note that introducing the $SemanticReranker$ (w/ SR) always outperforms the DPR+FiD baseline (w/o SR), and at the same time it allows for a 5-fold input size reduction (top-$10$) while obtaining on-par or better results. In addition, a further finetuning step of the FiD (w/ SR + FT) outperforms the results obtained by the $SemanticReranker$ (w/ SR) by 1.7 and 2.9 F1 points on \mbox{\textsc{TopiOCQA}} and \mbox{OR-QuAC}, respectively. Further experiments and ablation studies are provided in Appendix \ref{sec:appendix}.
\newline \indent Finally, in Figure \ref{fig:latency_vs_f1_score} it can also be observed that using top-$10$ instead of top-$50$ can reduce FiD's latency by 60\% on average across the two datasets. We conducted a latency measurement to evaluate the impact of the $SemanticReranker$ and its associated parameters, with detailed information available in Appendix \ref{sec:appendix}. Given that the $SemanticReranker$ consists of a single $TransformerEncoder$ layer, its parameters are negligible when compared to both the DPR and FiD. Moreover, the $SemanticReranker$ accounts only for 0.34\% of the overall latency of the FiD reader, adding an additional 2.4ms per example on top of the 710ms taken by FiD. It is important to note that this impact is only considered in relation to FiD, as the retrieval phase remains constant regardless of the inclusion of the $SemanticReranker$.

\section{Conclusions}
\label{sec:conclusions}

In this paper, we identified several limitations of the typical Depnse Passage Retrieval (DPR) retriever and Fusion-in-Decoder (FiD) reader pipeline when applied in an ODConvQA setting. We proposed and evaluated an improved approach by including a fast reranking component between these two modules and by performing targeted finetuning steps. The proposed R3FINE strategy lead to a better latency/performance trade-off. The new baseline has proven to be both strong and efficient when compared to previous baselines, thus making it suitable for future comparisons of new approaches.

\section*{Limitations}
\label{sec:limitations}

The study presented in this work aimed to identify and address various limitations of the commonly used ODConvQA pipeline. While our approach may not be technically groundbreaking, the work’s novelty lies in the presented findings to design strong and efficient baselines for ODConvQA. It should be noted that further research is needed to compare the performance of the proposed R3FINE strategy with other rerankers on non-conversational QA datasets, which would provide valuable insights into how effective the R3FINE approach is in different contexts.

\bibliography{anthology,custom}
\bibliographystyle{acl_natbib}

\appendix

\section{Appendix}
\label{sec:appendix}

This section provides additional information in support of the work done within the paper.

\subsection{\mbox{OR-QuAC} conversion to \mbox{\textsc{TopiOCQA}}'s format and models} 
\label{subsec:orquac_conversion_to_topiocqa}

To make the \mbox{OR-QuAC} dataset compatible with \mbox{\textsc{TopiOCQA}}'s models, we applied the following steps:
\begin{itemize}
    \item all answers of type \textsc{CANNOTANSWER} and \textsc{NOTRECOVERED} have been mapped to \textsc{UNANSWERABLE}
    \item for the DPR's $ConversationEncoder$ component, we followed the \mbox{\textsc{TopiOCQA}}'s \textsc{AllHistory} conversation representation
    \item for the DPR's $PassageEncoder$ component, each passage title has been reduced from \texttt{"passage\_page\_title [SEP] passage\_page\_subtitle"} to \texttt{"passage\_page\_title"}. This is due to the fact that, compared to \mbox{\textsc{TopiOCQA}}, \mbox{OR-QuAC} does not provide the information about the section where a particular passage is located within the page.
    \item for the FiD component, each passage information has been reduced from \texttt{"title: sub-title: context:"} to \texttt{"title: context:"}. This is done for the same reason mentioned in the previous point.
\end{itemize}
Each passage text in \mbox{OR-QuAC}'s Wikipedia knowledge source has been mapped to its corresponding embedding via the \mbox{\textsc{TopiOCQA}}'s $PassageEncoder$ component. We then performed the same retrieval step as the one done for \mbox{\textsc{TopiOCQA}}. We exploited \mbox{\textsc{TopiOCQA}}'s DPR module for both datasets as the retrieval phase is very similar between the two. However, given that, unlike \mbox{\textsc{TopiOCQA}}, \mbox{OR-QuAC} is of extractive type, we had to train the FiD module from scratch. We followed the same training configuration as the one used for \mbox{\textsc{TopiOCQA}}. 

\begin{table*}[t]
\centering
\begin{tabular}{c| c| c| c| c| c| c}
    \hline
    $L$ & $h_{c_1}$ & $h_{c_1}, \dots, h_{c_i}$ & $p_1, \dots, p_k$ & $CrossEncoder$ & \textbf{\mbox{\textsc{TopiOCQA}}} & \textbf{\mbox{OR-QuAC}} \\
    \hline
    1 & \checkmark & & \checkmark & \checkmark & 78.85 & 71.51 \\
    1 & & \checkmark & \checkmark & \checkmark & 78.88 & 71.48 \\
    4 & & \checkmark & \checkmark & \checkmark & 78.84 & 71.32 \\
    \hline
    1 & \checkmark & & \checkmark & & 77.17 & 69.83 \\
    1 & & \checkmark & \checkmark & & 77.35 & 70.18 \\
    4 & & \checkmark & \checkmark & & 77.30 & 70.45 \\
    \hline
    1 & & & \checkmark & & 67.07 & 64.18 \\
    \hline
\end{tabular}
\caption{Average top-$k$ results obtained on the \mbox{\textsc{TopiOCQA}} and \mbox{OR-QuAC} dev split, with different configurations.}
\label{tab:reranker_ablation_study}
\end{table*}

\subsection{Reranker training and ablation study} 
\label{subsec:reranker_training_and_ablation_study}

We tried different configurations of the $SemanticReranker$ to find the most efficient and effective one. In addition to the decision of whether to finetune the DPR's $ConversationEncoder$ together with the $SemanticReranker$, we also tried varying the number of layers $L$ of the $SemanticReranker$ from 1 to 4 and changing its input, by choosing a combination from:
\begin{itemize}
    \item $h_{c_1}$: use of conversation's history dense representation
    \item $h_{c_1}, \dots, h_{c_i}$: use of conversation's history tokens dense representation
    \item $p_1, \dots, p_k$: use of passages dense representation 
\end{itemize}

Table \ref{tab:reranker_ablation_study} shows the average top-$k$ results obtained on the \mbox{\textsc{TopiOCQA}} and \mbox{OR-QuAC} dev split, where $k$ varies between 1, 3, 5, 10, 15, 20, 30, 50, 100, 250, 500, 750, and 1000. For \mbox{\textsc{TopiOCQA}}, we report the presence of the gold passage within the top-$k$ limit. For \mbox{OR-QuAC} we report the presence of the gold answer within the top-$k$ limit.
\newline \indent Given that the MIPS function cannot be applied when $p_1, \dots, p_k$ representations are used alone, i.e, without the conversation history, we applied a linear projection on top of the $SemanticReranker$ to obtain a score for each passage in input.
\newline \indent As far as training the $SemanticReranker$ is concerned, we trained it for 10 epochs when finetuned together with the $ConversationEncoder$. We instead trained it for 20 epochs when the $ConversationEncoder$ was kept frozen and when $p_1, \dots, p_k$ representations were used alone. We leveraged the same objective function used for training the initial DPR. We used early stopping to chose the best performing model on the dev set. We also used a linear learning rate decay throughout the training process, and AdamW with a learning rate of 5e-5 and weight decay of 1e-2.
\newline \indent Among the different combinations shown in Table \ref{tab:reranker_ablation_study}, we considered the first entry as the best choice, i.e., the model with $L = 1$, $h_{c_1}$, $p_1, \dots, p_k$, and $CrossEncoder$ finetuning.

\subsection{Retriever results}
\label{subsec:retriever_results}

Table \ref{tab:topiocqa_orquac_train_coverage_full_table} shows the train split retrieval coverage before/after the introduction of the $SemanticReranker$ (w/o SR) when a larger number of passages is considered (50 vs 1000). For both datasets, we report the presence of the gold passage within the top-$k$ limit.

\begin{table}[h]
\centering
\begin{tabular}{c| c c| c c}
    \hline
    \multicolumn{1}{c|}{} & \multicolumn{2}{c|}{\textbf{\mbox{\textsc{TopiOCQA}}}} & \multicolumn{2}{c}{\textbf{\mbox{OR-QuAC}}} \\
    \hline
    \multicolumn{1}{c|}{\textbf{top-$k$}} & \multicolumn{1}{c|}{\textbf{w/o SR}} & \multicolumn{1}{c|}{\textbf{w/ SR}} & \multicolumn{1}{c|}{\textbf{w/o SR}} & \multicolumn{1}{c}{\textbf{w/ SR}} \\
    \hline
    1 & 31.54 & 98.25 & 31.69 & 77.31 \\
    5 & 66.99 & 99.71 & 59.08 & 89.10 \\
    10 & 78.29 & 99.72 & 66.31 & 89.36 \\
    20 & 86.78 & 99.72 & 71.77 & 89.40 \\
    50 & 93.55 & 99.72 & 77.35 & 89.40 \\
    500 & 99.36 & 99.72 & 87.15 & 89.41 \\
    \hline
    1000 & 99.72 & 99.72 & 89.41 & 89.41 \\
    \hline
\end{tabular}
\caption{Train split retrieval coverage before/after the introduction of the $SemanticReranker$ (w/o SR) when a larger number of passages is considered (50 vs 1000). For both datasets, we report the presence of the gold passage within the top-$k$ limit.}
\label{tab:topiocqa_orquac_train_coverage_full_table}
\end{table}

\begin{table}[h]
\centering
\begin{tabular}{c| c c| c c}
    \hline
    \multicolumn{1}{c|}{} & \multicolumn{2}{c|}{\textbf{\mbox{\textsc{TopiOCQA}}}} & \multicolumn{2}{c}{\textbf{\mbox{OR-QuAC}}} \\
    \hline
    \multicolumn{1}{c|}{\textbf{top-$k$}} & \multicolumn{1}{c|}{\textbf{w/o SR}} & \multicolumn{1}{c|}{\textbf{w/ SR}} & \multicolumn{1}{c|}{\textbf{w/o SR}} & \multicolumn{1}{c}{\textbf{w/ SR}} \\
    \hline
    1 & 24.66 & 42.64 & 29.27 & 56.64 \\
    5 & 51.87 & 68.62 & 51.63 & 70.82 \\
    10 & 62.45 & 75.62 & 57.14 & 72.27 \\
    20 & 70.21 & 80.75 & 61.37 & 72.80 \\
    50 & 77.41 & 84.69 & 65.22 & 73.53 \\
    500 & 89.58 & 90.73 & 72.97 & 74.34 \\
    \hline
    1000 & 91.49 & 91.49 & 74.55 & 74.55 \\
    \hline
\end{tabular}
\caption{Dev split retrieval coverage before/after the introduction of the $SemanticReranker$ (w/o SR) when a larger number of passages is considered (50 vs 1000). For \mbox{\textsc{TopiOCQA}}, we report the presence of the gold passage within the top-$k$ limit. For \mbox{OR-QuAC} we report the presence of the gold answer within the top-$k$ limit.}
\label{tab:topiocqa_orquac_dev_coverage_full_table}
\end{table}

Table \ref{tab:topiocqa_orquac_dev_coverage_full_table} shows the dev split retrieval coverage before/after the introduction of the $SemanticReranker$ (w/o SR) when a larger number of passages is considered (50 vs 1000). For \mbox{\textsc{TopiOCQA}}, we report the presence of the gold passage within the top-$k$ limit. For \mbox{OR-QuAC} we report the presence of the gold answer within the top-$k$ limit. Table \ref{tab:orquac_test_coverage_full_table} shows the \mbox{OR-QuAC} test split retrieval coverage before/after the introduction of the $SemanticReranker$ (w/o SR) when a larger number of passages is considered (50 vs 1000). We report the presence of the gold answer within the top-$k$ limit.

\begin{table}[!h]
\centering
\begin{tabular}{c| c c}
    \hline
    \multicolumn{1}{c|}{} & \multicolumn{2}{c}{\textbf{\mbox{OR-QuAC}}} \\
    \hline
    \multicolumn{1}{c|}{\textbf{top-$k$}} & \multicolumn{1}{c|}{\textbf{w/o SR}} & \multicolumn{1}{c}{\textbf{w/ SR}} \\
    \hline
    1 & 26.82 & 48.75 \\
    5 & 46.29 & 64.71 \\
    10 & 51.50 & 66.47 \\
    20 & 55.63 & 67.31 \\
    50 & 59.85 & 68.10 \\
    500 & 67.87 & 69.48 \\
    \hline
    1000 & 69.86 & 69.86 \\
    \hline
\end{tabular}
\caption{\mbox{OR-QuAC} test split retrieval coverage before/after the introduction of the $SemanticReranker$ (w/o SR) when a larger number of passages is considered (50 vs 1000). We report the presence of the gold answer within the top-$k$ limit.}
\label{tab:orquac_test_coverage_full_table}
\end{table}

\subsection{Reader results}
\label{subsec:reader_results}

Table \ref{tab:topiocqa_fid_dev_results_full_table}, Table \ref{tab:orquac_fid_dev_results_full_table}, and Table \ref{tab:orquac_fid_test_results_full_table} show the impact the introduction of the $SemanticReranker$ has on the FiD reader. The input to the FiD reader are either passages returned by the initial DPR retriever (w/o SR) or passages returned by the $SemanticReranker$ (w/ SR). 

\begin{table}[!h]
\centering
\begin{tabular}{c| c c| c c}
    \hline
    \multicolumn{5}{c}{\textbf{\mbox{\textsc{TopiOCQA}}}} \\
    \hline
    {} & \multicolumn{2}{c|}{\textbf{w/o SR}} & \multicolumn{2}{c}{\textbf{w/ SR}} \\
    \hline
    \multicolumn{1}{c|}{\textbf{top-$k$}} & \multicolumn{1}{c|}{\textbf{EM}} & \multicolumn{1}{c|}{\textbf{F1}} & \multicolumn{1}{c|}{\textbf{EM}} & \multicolumn{1}{c}{\textbf{F1}} \\
    \hline
    1 & 19.3 & 37.6 & 28.1 & 50.4 \\
    5 & 27.0 & 49.6 & 32.2 & 56.5 \\
    10 & 29.8 & 52.4 & 33.2 & 57.3 \\
    20 & 31.3 & 54.0 & 33.4 & 56.5 \\
    50 & 33.0 & 55.1 & 33.9 & 56.2 \\
    \hline
\end{tabular}
\caption{FiD reader performance (Exact Match and F1 scores) on the \mbox{\textsc{TopiOCQA}} dev split before/after the introduction of the $SemanticReranker$ (w/o SR).}
\label{tab:topiocqa_fid_dev_results_full_table}
\end{table}

\begin{table}[!h]
\centering
\begin{tabular}{c| c c| c c}
    \hline
    \multicolumn{5}{c}{\textbf{\mbox{OR-QuAC}}} \\
    \hline
    {} & \multicolumn{2}{c|}{\textbf{w/o SR}} & \multicolumn{2}{c}{\textbf{w/ SR}} \\
    \hline
    \multicolumn{1}{c|}{\textbf{top-$k$}} & \multicolumn{1}{c|}{\textbf{EM}} & \multicolumn{1}{c|}{\textbf{F1}} & \multicolumn{1}{c|}{\textbf{EM}} & \multicolumn{1}{c}{\textbf{F1}} \\
    \hline
    1 & 13.2 & 22.2 & 12.2 & 25.9 \\
    5 & 16.4 & 26.3 & 16.4 & 28.7 \\
    10 & 18.0 & 27.6 & 18.0 & 29.4 \\
    20 & 19.2 & 28.2 & 18.9 & 29.2 \\
    50 & 19.6 & 27.7 & 19.3 & 27.7 \\
    \hline
\end{tabular}
\caption{FiD reader performance (Exact Match and F1 scores) on the \mbox{OR-QuAC} dev split before/after the introduction of the $SemanticReranker$ (w/o SR).}
\label{tab:orquac_fid_dev_results_full_table}
\end{table}

\begin{table}[!h]
\centering
\begin{tabular}{c| c c| c c}
    \hline
    \multicolumn{5}{c}{\textbf{\mbox{OR-QuAC}}} \\
    \hline
    {} & \multicolumn{2}{c|}{\textbf{w/o SR}} & \multicolumn{2}{c}{\textbf{w/ SR}} \\
    \hline
    \multicolumn{1}{c|}{\textbf{top-$k$}} & \multicolumn{1}{c|}{\textbf{EM}} & \multicolumn{1}{c|}{\textbf{F1}} & \multicolumn{1}{c|}{\textbf{EM}} & \multicolumn{1}{c}{\textbf{F1}} \\
    \hline
    1 & 14.7 & 23.0 & 13.9 & 25.7 \\
    5 & 17.8 & 27.1 & 17.9 & 29.5 \\
    10 & 19.0 & 28.2 & 19.4 & 30.0 \\
    20 & 20.3 & 29.1 & 20.8 & 30.5 \\
    50 & 22.0 & 29.9 & 22.1 & 30.2 \\
    \hline
\end{tabular}
\caption{FiD reader performance (Exact Match and F1 scores) on the \mbox{OR-QuAC} test split before/after the introduction of the $SemanticReranker$ (w/o SR).}
\label{tab:orquac_fid_test_results_full_table}
\end{table}

\subsection{Reader is susceptible to noisy input}
\label{subsec:reader_susceptibility_to_noisy_input}

Table \ref{tab:topiocqa_performance_full_table} shows the FiD reader performance on the \mbox{\textsc{TopiOCQA}} dev split, with/without the gold passage (w/o gold) in the top-$k$ limit.  This analysis is limited to the \mbox{\textsc{TopiOCQA}} dataset as it is the only one to provide information about the gold passage for the dev split.

\begin{table}[!h]
\centering
\begin{tabular}{c| c c| c c}
    \hline
    \multicolumn{5}{c}{\textbf{\mbox{\textsc{TopiOCQA}}}} \\
    \hline
    {} & \multicolumn{2}{c|}{\textbf{w/o gold}} & \multicolumn{2}{c}{\textbf{w/ gold}} \\
    \hline
    \multicolumn{1}{c|}{\textbf{top-$k$}} & \multicolumn{1}{c|}{\textbf{EM}} & \multicolumn{1}{c|}{\textbf{F1}} & \multicolumn{1}{c|}{\textbf{EM}} & \multicolumn{1}{c}{\textbf{F1}} \\
    \hline
    1 & 19.3 & 37.6 & 38.3 & 65.5 \\
    5 & 27.0 & 49.6 & 36.3 & 62.5 \\
    10 & 29.8 & 52.4 & 35.8 & 61.5 \\
    20 & 31.3 & 54.0 & 36.2 & 60.8 \\
    50 & 33.0 & 55.1 & 35.9 & 59.5  \\
    \hline
\end{tabular}
\caption{FiD reader performance (Exact Match and F1 scores) on the \mbox{\textsc{TopiOCQA}} dev split, with/without the gold passage (w/o gold) in the top-$k$ limit.}
\label{tab:topiocqa_performance_full_table}
\end{table}

\subsection{Further reader study} 
\label{subsec:fid_reader_finetuning}

To better understand the impact that the introduction of the $SemanticReranker$ has on the FiD reader, Table \ref{tab:topiocqa_non_finetuned_fid_plus_trained_fid_dev_results}, Table \ref{tab:orquac_non_finetuned_fid_plus_trained_fid_dev_results}, and Table \ref{tab:orquac_non_finetuned_fid_plus_trained_fid_test_results} show the results obtained after taking a non-finetuned FiD and training it on the top-$10$ passages returned by the initial DPR retriever and on the top-$10$ passages returned by the $SemanticReranker$. On both datasets, we followed the same training configuration as the one used for \mbox{\textsc{TopiOCQA}}.

\begin{table}[!h]
\centering
\begin{tabular}{c| c c| c c}
    \hline
    \multicolumn{5}{c}{\textbf{\mbox{\textsc{TopiOCQA}}}} \\
    \hline
    {} & \multicolumn{2}{c|}{\textbf{w/o SR}} & \multicolumn{2}{c}{\textbf{w/ SR}} \\
    \hline
    \multicolumn{1}{c|}{\textbf{top-$k$}} & \multicolumn{1}{c|}{\textbf{EM}} & \multicolumn{1}{c|}{\textbf{F1}} & \multicolumn{1}{c|}{\textbf{EM}} & \multicolumn{1}{c}{\textbf{F1}} \\
    \hline
    1 & 19.2 & 37.8 & 29.5 & 50.9 \\
    5 & 27.8 & 49.9 & 34.0 & 56.9 \\
    10 & 30.4 & 52.0 & 34.3 & 57.0 \\
    \hline
\end{tabular}
\caption{FiD reader performance (Exact Match and F1 scores) on the \mbox{\textsc{TopiOCQA}} dev split after taking a non-finetuned FiD and training it on the top-$10$ passages returned by the initial DPR retriever (w/o SR) and on the top-$10$ passages returned by the $SemanticReranker$ (w/ SR).}
\label{tab:topiocqa_non_finetuned_fid_plus_trained_fid_dev_results}
\end{table}

\begin{table}[!h]
\centering
\begin{tabular}{c| c c| c c}
    \hline
    \multicolumn{5}{c}{\textbf{\mbox{OR-QuAC}}} \\
    \hline
    {} & \multicolumn{2}{c|}{\textbf{w/o SR}} & \multicolumn{2}{c}{\textbf{w/ SR}} \\
    \hline
    \multicolumn{1}{c|}{\textbf{top-$k$}} & \multicolumn{1}{c|}{\textbf{EM}} & \multicolumn{1}{c|}{\textbf{F1}} & \multicolumn{1}{c|}{\textbf{EM}} & \multicolumn{1}{c}{\textbf{F1}} \\
    \hline
    1 & 14.2 & 23.1 & 13.9 & 27.6 \\
    5 & 18.0 & 27.7 & 18.1 & 30.8 \\
    10 & 19.2 & 28.9 & 19.3 & 31.2 \\
    \hline
\end{tabular}
\caption{FiD reader performance (Exact Match and F1 scores) on the \mbox{OR-QuAC} dev split after taking a non-finetuned FiD and training it on the top-$10$ returned by the initial DPR retriever (w/o SR) and on the top-$10$ returned by the $SemanticReranker$ (w/ SR).}
\label{tab:orquac_non_finetuned_fid_plus_trained_fid_dev_results}
\end{table}

\begin{table}[!h]
\centering
\begin{tabular}{c| c c| c c}
    \hline
    \multicolumn{5}{c}{\textbf{\mbox{OR-QuAC}}} \\
    \hline
    {} & \multicolumn{2}{c|}{\textbf{w/o SR}} & \multicolumn{2}{c}{\textbf{w/ SR}} \\
    \hline
    \multicolumn{1}{c|}{\textbf{top-$k$}} & \multicolumn{1}{c|}{\textbf{EM}} & \multicolumn{1}{c|}{\textbf{F1}} & \multicolumn{1}{c|}{\textbf{EM}} & \multicolumn{1}{c}{\textbf{F1}} \\
    \hline
    1 & 15.5 & 23.9 & 15.8 & 28.3 \\
    5 & 19.6 & 29.1 & 19.4 & 31.7 \\
    10 & 20.9 & 29.7 & 20.6 & 31.9 \\
    \hline
\end{tabular}
\caption{FiD reader performance (Exact Match and F1 scores) on the \mbox{OR-QuAC} test split after taking a non-finetuned FiD and training it on the top-$10$ passages returned by the initial DPR retriever (w/o SR) and on the top-$10$ passages returned by the $SemanticReranker$ (w/ SR).}
\label{tab:orquac_non_finetuned_fid_plus_trained_fid_test_results}
\end{table}

Table \ref{tab:topiocqa_finetuned_fid_plus_finetuned_fid_dev_results}, Table \ref{tab:orquac_finetuned_fid_plus_finetuned_fid_dev_results}, and Table \ref{tab:orquac_finetuned_fid_plus_finetuned_fid_test_results} show instead the results obtained after taking an already finetuned FiD reader and further finetuning it on the top-$10$ passages returned by the initial DPR retriever and on the top-$10$ passages returned by the $SemanticReranker$. On both datasets, the amount of finetuning steps is equal to the one used for training the already finetuned FiD reader.

\begin{table}[!h]
\centering
\begin{tabular}{c| c c| c c}
    \hline
    \multicolumn{5}{c}{\textbf{\mbox{\textsc{TopiOCQA}}}} \\
    \hline
    {} & \multicolumn{2}{c|}{\textbf{w/o SR}} & \multicolumn{2}{c}{\textbf{w/ SR}} \\
    \hline
    \multicolumn{1}{c|}{\textbf{top-$k$}} & \multicolumn{1}{c|}{\textbf{EM}} & \multicolumn{1}{c|}{\textbf{F1}} & \multicolumn{1}{c|}{\textbf{EM}} & \multicolumn{1}{c}{\textbf{F1}} \\
    \hline
    1 & 21.5 & 39.3 & 30.7 & 52.4 \\
    5 & 30.4 & 52.3 & 36.1 & 59.4 \\
    10 & 32.8 & 54.6 & 35.8 & 59.0 \\
    \hline
\end{tabular}
\caption{FiD reader performance (Exact Match and F1 scores) on the \mbox{\textsc{TopiOCQA}} dev split after taking an already finetuned FiD and further finetuning it on the top-$10$ returned by the initial DPR retriever (w/o SR) and on the top-$10$ returned by the $SemanticReranker$ (w/ SR).}
\label{tab:topiocqa_finetuned_fid_plus_finetuned_fid_dev_results}
\end{table}

\begin{table}[!h]
\centering
\begin{tabular}{c| c c| c c}
    \hline
    \multicolumn{5}{c}{\textbf{\mbox{OR-QuAC}}} \\
    \hline
    {} & \multicolumn{2}{c|}{\textbf{w/o SR}} & \multicolumn{2}{c}{\textbf{w/ SR}} \\
    \hline
    \multicolumn{1}{c|}{\textbf{top-$k$}} & \multicolumn{1}{c|}{\textbf{EM}} & \multicolumn{1}{c|}{\textbf{F1}} & \multicolumn{1}{c|}{\textbf{EM}} & \multicolumn{1}{c}{\textbf{F1}} \\
    \hline
    1 & 15.2 & 23.8 & 15.4 & 29.0 \\
    5 & 18.9 & 28.4 & 18.4 & 31.2 \\
    10 & 19.9 & 29.4 & 20.1 & 32.0 \\
    \hline
\end{tabular}
\caption{FiD reader performance (Exact Match and F1 scores) on the \mbox{OR-QuAC} dev split after taking an already finetuned FiD and further finetuning it on the top-$10$ passages returned by the initial DPR retriever (w/o SR) and on the top-$10$ passages returned by the $SemanticReranker$ (w/ SR).}
\label{tab:orquac_finetuned_fid_plus_finetuned_fid_dev_results}
\end{table}

\begin{table}[!h]
\centering
\begin{tabular}{c| c c| c c}
    \hline
    \multicolumn{5}{c}{\textbf{\mbox{OR-QuAC}}} \\
    \hline
    {} & \multicolumn{2}{c|}{\textbf{w/o SR}} & \multicolumn{2}{c}{\textbf{w/ SR}} \\
    \hline
    \multicolumn{1}{c|}{\textbf{top-$k$}} & \multicolumn{1}{c|}{\textbf{EM}} & \multicolumn{1}{c|}{\textbf{F1}} & \multicolumn{1}{c|}{\textbf{EM}} & \multicolumn{1}{c}{\textbf{F1}} \\
    \hline
    1 & 16.7 & 24.9 & 16.6 & 28.9 \\
    5 & 20.0 & 29.2 & 20.1 & 32.2 \\
    10 & 21.4 & 30.0 & 21.6 & 32.9 \\
    \hline
\end{tabular}
\caption{FiD reader performance (Exact Match and F1 scores) on the \mbox{OR-QuAC} test split after taking an already finetuned FiD and further finetuning it on the top-$10$ passages returned by the initial DPR retriever (w/o SR) and on the top-$10$ passages returned by the $SemanticReranker$ (w/ SR).}
\label{tab:orquac_finetuned_fid_plus_finetuned_fid_test_results}
\end{table}

\subsection{Latency measurement} 
\label{subsec:latency_measurement}

Latency measurement (see Figure \ref{fig:latency_vs_f1_score}) has been performed on the same NVIDIA V100 16GB GPU, by following the FiD's \texttt{test\_reader.py} script provided with the \mbox{\textsc{TopiOCQA}} dataset. We set the \texttt{per\_gpu\_batch\_size} paramenter to 4 in all runs and chose the value of the \texttt{n\_context} parameter from 1, 3, 5, 10, 20, 30, and 50, based on the number of input passages. For each value, we report the latency relative to the maximum \texttt{n\_context} parameter value, i.e., 50. We used CUDA events synchronization markers to measure the elapsed time for the preprocessing and evaluation of \mbox{\textsc{TopiOCQA}}'s dev split and \mbox{OR-QuAC}'s test split.

\end{document}